\documentclass[runningheads]{llncs}

\usepackage{graphicx}
\usepackage{paralist}
\usepackage{todonotes}
\usepackage{acronym}
\usepackage{bm}
\usepackage{amssymb}
\usepackage{graphicx}
\usepackage{listings}
\usepackage{float}
\usepackage{multirow}
\usepackage[hyphens]{url}

\newcommand\blfootnote[1]{%
  \begingroup
  \renewcommand\thefootnote{}\footnote{#1}%
  \addtocounter{footnote}{-1}%
  \endgroup
}

\begin{document}

\title{The KEEN Universe}
\subtitle{An Ecosystem for Knowledge Graph Embeddings with a Focus on Reproducibility and Transferability\blfootnote{This paper has been published under Springer for the International Semantic Web Conference}}

\author{Mehdi Ali\inst{1,2} \and
Hajira Jabeen\inst{1} \and
Charles Tapley Hoyt \inst{3} \and
Jens Lehmann\inst{1,2}}
\authorrunning{Ali \textit{et al.}}

\institute{Smart Data Analytics Group, University of Bonn, Germany \\
\email{\{mehdi.ali,jabeen,jens.lehmann\}@cs.uni-bonn.de} \and
Department of Enterprise Information Systems, Fraunhofer Institute for Intelligent Analysis and
Information Systems (IAIS), Sankt Augustin and Dresden, Germany\\
\email{\{mehdi.ali,jens.lehmann\}@iais.fraunhofer.de}\and
Department of Bioinformatics, Fraunhofer Institute for Algorithms and Scientific Computing (SCAI), Sankt Augustin, Germany\\
\email{charles.hoyt@scai.fraunhofer.de}
}

\maketitle

\begin{abstract}
There is an emerging trend of embedding knowledge graphs (KGs) in continuous vector spaces in order to use those for machine learning tasks. Recently, many knowledge graph embedding (KGE) models have been proposed that learn low dimensional representations while trying to maintain the structural properties of the KGs such as the similarity of nodes depending on their edges to other nodes. KGEs can be used to
address tasks within KGs such as the prediction of novel links and the disambiguation of entities. They can also be used for downstream tasks like question answering and fact-checking. Overall, these tasks are relevant for the semantic web community. Despite their popularity, the reproducibility of KGE experiments and the transferability of proposed KGE models to research fields outside the machine learning community can be a major challenge. Therefore, we present the KEEN Universe, an ecosystem for knowledge graph embeddings that we have developed with a strong focus on reproducibility and transferability. The KEEN Universe currently consists of the Python packages PyKEEN (Python KnowlEdge EmbeddiNgs), BioKEEN (Biological KnowlEdge EmbeddiNgs), and the KEEN Model Zoo for sharing trained KGE models with the community.

\vspace{0.2cm}
\textbf{Resource Type}: Software Framework
\newline
\textbf{License:} MIT License
\newline
\textbf{Permanent URL}: \url{https://figshare.com/articles/The_KEEN_Universe/7957445}
\end{abstract}

\section{Introduction}
In the last two decades, representing factual information as knowledge graphs (KGs) has gained significant attention. KGs have been successfully applied to tasks such as link prediction, clustering, and question answering. In the context of this paper, a KG is a directed, multi-relational graph that represents entities as nodes, and their relations as edges, and can be used as an abstraction of the real world. Factual information contained in KGs is represented as triples of the form \begin{math}(h,r,t)\end{math}, where $h$ and $t$ denote the head and tail entities, and $r$ denotes their respective relation.
Prominent examples of KGs are DBpedia~\cite{swj_dbpedia}, Wikidata~\cite{vrandevcic2014wikidata}, Freebase~\cite{bollacker2008freebase}, and Knowledge Vault~\cite{dong2014knowledge}. Traditionally, KGs have been processed in their essential form as symbolic systems, but recently, knowledge graph embedding models (KGEs) have become popular that encode the nodes and edges of KGs into low-dimensional continuous vector spaces while best preserving the structural properties of the KGs. The learned embeddings can be used to perform algebraic operations on the corresponding KGs, and common tasks are link prediction and entity disambiguation~\cite{wang2017knowledge}. Furthermore, we can observe that KGEs are applied in downstream tasks such as question answering (QA)~\cite{saha2018complex}.

Although KGEs are becoming popular, the reproducibility of KGE experiments and the transferability of the proposed models to research fields outside the machine learning community such as the semantic web or the biomedical domain remains a challenge. Depending on the used hyper-parameter values and the optimization approach, the model performance can vary significantly. For instance, in the experiments performed by Akrami \textit{et al.}~\cite{Akrami:2018:REK:3269206.3269266} an increase of 14.4\% for the TransE model and 23.6\% for the DistMult model in the \textit{hits@k} metric has been reported. However, the reasons for the performance discrepancies are often not discussed in depth~\cite{xiao2017ssp,xie2016representation}, impeding the reproducibility of experiments. 
Furthermore, applying proposed KGE models requires both expertise in KGEs and in implementing these models which can be obstacles for non-machine learning researchers. These are significant shortcomings considering that in research fields like the semantic web or the bioinformatics community, KGs are widely applied, and KGE models might have a strong potential to be used in many tasks.
Initiatives like the SIGMOD\footnote{http://db-reproducibility.seas.harvard.edu/} guidelines defined by the database community or the FAIR data principles~\cite{wilkinson2016fair} highlight that reproducibility and transferability is not only a fundamental challenge inside the research field of KGEs, but it is a cross-domain issue.

In this paper, we describe a software ecosystem that we have developed with a strong emphasis on reproducibility and transferability. Our contribution is the KEEN Universe that currently consists of: i.) PyKEEN (Python KnowlEdge Graph EmbeddiNgs), a Python package encapsulating the machine learning functionalities, ii.) BioKEEN (Biological KnowlEdge Graph EmbeddiNgs)~\cite{ali2019biokeen}, a Python package specifically developed to facilitate the use of KGEs within the bioinformatics community and iii.) the KEEN Model Zoo, a platform to share pre-trained KGE models. Furthermore, we evaluate the usability of the KEEN Universe on two case scenarios from the area of scholarly metadata research and bioinformatics.

\section{Impact and Use Cases}
\subsection{Impact}
\paragraph{\textbf{Impact on the KGE Community.}} By providing an ecosystem that enables researchers to easily share code, experimental set-ups and research results without requiring additional overhead, the KEEN Universe is an essential step in the direction of reproducible KGE research. Specifically, researchers can integrate their new KGE models into our ecosystem to enhance comparability with existing approaches as well as to share their trained models through our model zoo to make it easily accessible for the community. The functionalities provided by the KEEN Universe will save researchers significant amount of time and facilitate the work on complex tasks.

\paragraph{\textbf{Impact Beyond the KGE Community.}} KGs have become a standard in representing factual information across different domains. Considering that KGs are often incomplete and noisy, the KEEN Universe can be applied in numerous applications to derive new facts. For instance, the KEEN Universe has been used on scholarly KGs to provide research recommendations~\cite{KGErecommender} and on biomedical KGs to predict associations between biomedical entities~\cite{ali2019biokeen,kulmanov2019embeddings}. Moreover, it can be used in downstream tasks like QA and dialogue generation~\cite{bonatti2019knowledge,nickel2016review}. 

\paragraph{\textbf{Impact on Industry.}} KGs are established in several major companies such as Google, Facebook, Bayer, Siemens, and KGEs are for instance used to build KGE based recommender systems~\cite{bonatti2019knowledge,hildebrandt2018configuration}. Furthermore, the evolution of industry to \textit{Industry 4.0} paves a new way for KGEs to be applied in the observation of manufacturing processes: (knowledge) graphs are a convenient approach to model the data produced by sensors which can be used to model the status of production pipelines. The encoded information can be fed to machine learning based systems for predictive maintenance. Instead of performing feature engineering which is time-consuming and complex, KGEs can be used to encode the information of KGs~\cite{garofalo2018leveraging}. Enterprises could use the KEEN Universe to experiment with KGEs before performing major investments to build their own specialized systems. 

\paragraph{\textbf{Impact on Teaching.}}The KEEN Universe can be used by students to learn how KGE models and their training and evaluation procedures are implemented which helps them to implement new KGE models that in turn could be integrated into the KEEN Universe. It has been already successfully applied in two master theses and currently, it is being used in a further master thesis to compare link prediction approaches based on handcrafted KG features against KGEs based link prediction approaches. Furthermore, it is used in the Knowledge Graph Analysis Lab (University of Bonn) to introduce KGE models to master students.

\subsection{Use Cases}
\paragraph{\textbf{Bioinformatics.}} Bio2Vec\footnote{http://bio2vec.net/} is a project that aims to provide a platform to enable the development of machine learning and data analytic tools for biological KGs with the goal of discovering molecular mechanisms underlying complex diseases and drugs’ modes of action. This project also aims to provide pre-trained embeddings for existing biological data, and additional data created and produced within this project. BioKEEN and PyKEEN have been applied already within Bio2Vec to predict hierarchies and cross-talks between biological pathways~\cite{ali2019biokeen} and to predict protein-protein interactions~\cite{kulmanov2019embeddings}. Furthermore, the model to predict interactions between biological pathways has been shared through the KEEN Model Zoo (\url{https://github.com/SmartDataAnalytics/KEEN-Model-Zoo/tree/master/bioinformatics/ComPath/compath_model_01}).

\paragraph{\textbf{Bayer Crop Science R\&D.}} The department of Computational Life Science (CLS) at Bayer Crop Science R\&D\footnote{https://agrar.bayer.de/}
developed a large knowledge graph to describe field trial experiments in which
candidates for crop protection products are tested across many experimental
settings. The knowledge graph is augmented with trial properties, wherein each node
contains information beyond the graph structure. However, a subgraph of the property
graph can be extracted in such a way that only important relationships are preserved
between nodes. This subgraph is stored as a collection of subject-predicate-object
triples to allow for a range of embedding techniques to be easily applied. Since
different use cases may require a different approach to mining the graph structure
for suggested links or node similarities, it is necessary to have a framework that
can simply consume the same graph data and apply new models without a large time
investment. 

The modular design of PyKEEN makes it a perfect fit for the needs of Bayer CLS
researchers. The knowledge graph contains nodes of various categories and relation
types, as well as many-to-one and one-to-many relations, requiring the use of
advanced embedding methods. In addition, new embedding algorithms can be simply
added to or modified from the existing framework. As an initial use case, Bayer CLS
researchers implemented the included TransR embedding method to their subgraph and,
with very little effort, produced an embedding space that demonstrated clear
clusters between node categories. Additionally, they were easily able to add node
category support to PyKEEN in order to extend the functionality of the existing
TransD algorithm. The team at Bayer CLS expects to provide insights into field trial
design, future field trial planning, and data quality checks using link predictions
from graph embeddings trained and optimized within PyKEEN.

\section{System Description}
To improve the reproducibility of KGE experiments, we have defined the following requirements for our ecosystem: i.) provide users the full control of the experimental setup, ii.) provide transparent training procedure for all KGE models, and iii.) provide identical evaluation procedure for all KGE models. To enable the transferability of KGE research, we have defined two requirements: i.) enable experts and inexperienced users to use the ecosystem, ii.) easy to specialize for requirements in different domains.
In the following, we explain how these requirements are addressed within the KEEN Universe. First, we describe PyKEEN (Section \ref{PyKEEN}), then we introduce BioKEEN (Section \ref{BioKEEN}), and finally, we present the KEEN Model Zoo (Section \ref{Model-Zoo}).

\subsection{PyKEEN}\label{PyKEEN}
Here, we present PyKEEN's software architecture, give an overview of the supported data formats, explain our approach for configuring KGE experiments, describe the training and evaluation procedures, describe which experimental artifacts are exported and finally, we present our inference workflow.

\begin{figure}[t]
\centering
\includegraphics[width=0.8\textwidth]{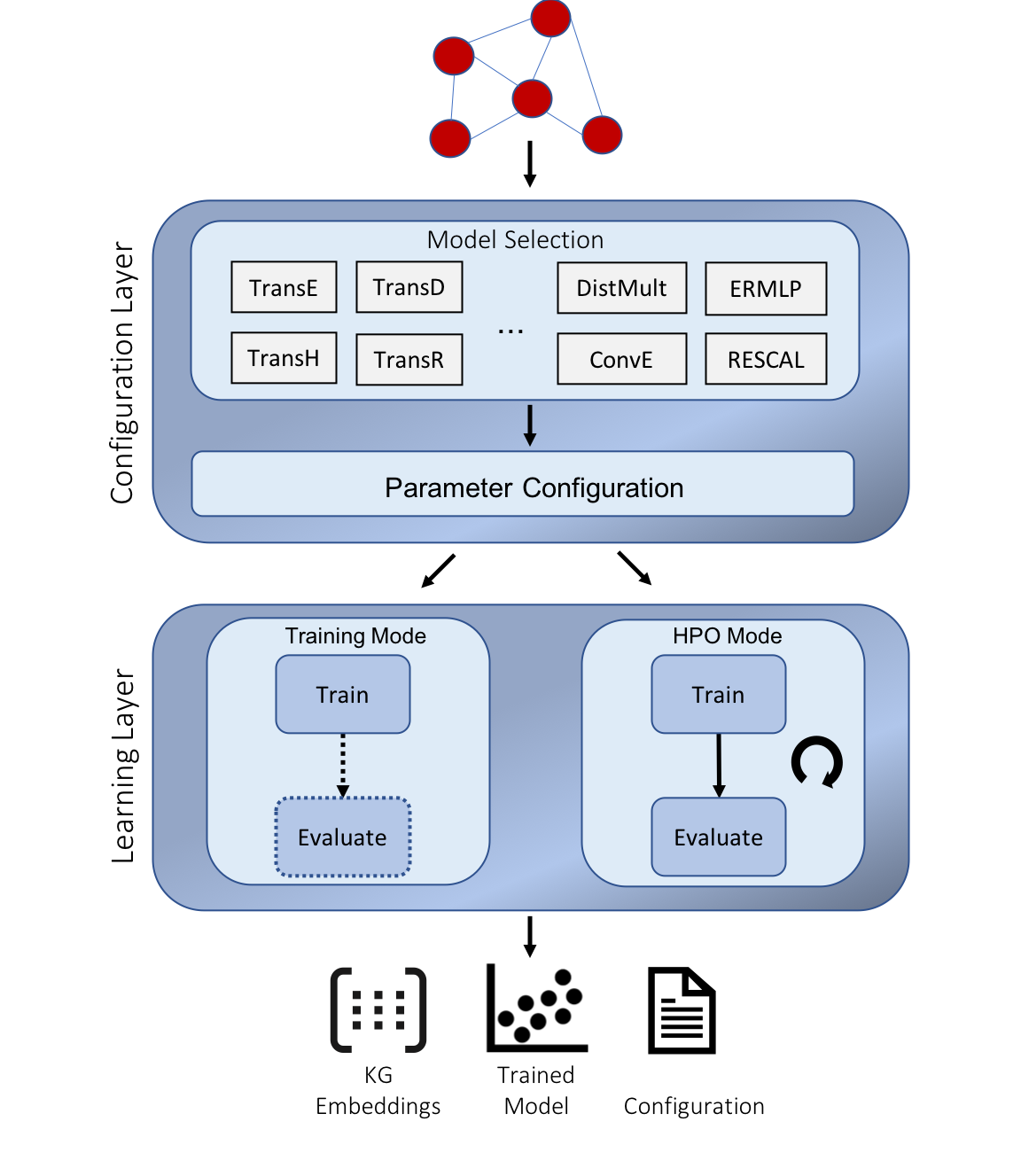}
\caption{Software architecture of PyKEEN: 1) the configuration layer assists users to specify experiments and 2) the learning layer trains a model with user-defined hyper-parameters or performs a hyper-parameter search.} 
\label{fig:pykeen_architecture}
\end{figure}

\paragraph{\textbf{Software Architecture.}}PyKEEN consists of a \textit{configuration} and a \textit{learning layer} (Figure \ref{fig:pykeen_architecture}). In the configuration layer, users can define their experiments, i.e. select the KGE model, its hyper-parameters, and define the evaluation procedure. The experimental setup is saved and passed to the learning layer that executes the experiment. In PyKEEN, a KGE model can be trained based on user defined hyper-parameter values or a hyper-parameter optimization can be performed to find suitable values. Finally, the experimental artifacts are exported.

\begin{figure}[t]
\centering
\includegraphics[width=0.9\textwidth]{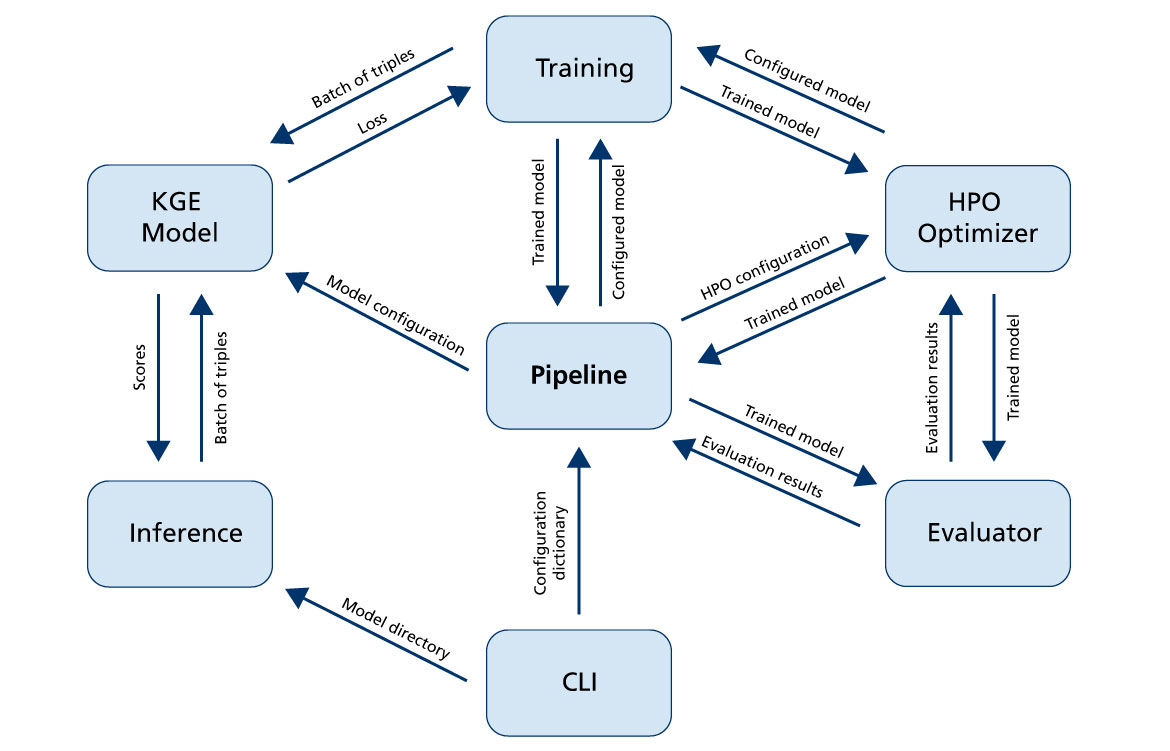}
\caption{PyKEEN's modules and their interactions~\cite{ali2019biokeen}.}
\label{fig:pykeen_modules}
\end{figure}

PyKEEN has a modular architecture (Figure \ref{fig:pykeen_modules}) and depending on the task different modules are executed and interact with each other. The \textit{command line interface (CLI)} module enables users to configure experiments through a terminal, the \textit{Pipeline} module starts and controls the configured experiment, \textit{KGEModel} modules represent KGE models, the \textit{Training} module is responsible for training a \textit{KGEModel} module and the \textit{Evaluator} module for its evaluation. A \textit{HPOOptimizer} module performs the hyper-parameter optimization (currently only \textit{random search} is available). To perform inference the \textit{Inference} module has been developed.

\paragraph{\textbf{Supported Data Formats.}}\label{supported_data_formats}
PyKEEN supports KGs represented as RDF, from NDEx~\cite{pratt2015ndex}, and as tab-separated values. We provide support for RDF, because it is an established data format to represent KGs~\cite{nickel2016review}. Examples of popular KGs available as RDF are DBpedia~\cite{swj_dbpedia} and Bio2RDF~\cite{belleau2008bio2rdf}. NDEx is an online commons for exchanging biological networks, and of interest for life science researchers. Finally, a tab separated file containing the triples of a KG can also be provided directly to PyKEEN. Overall, by supporting these data formats, many KGs can directly be used, allowing users to focus on their experiments rather than on data pre-processing.

\paragraph{\textbf{Configuration of Experiments.}} To provide users full control of the experimental setup we have developed the configuration layer (Figure \ref{fig:pykeen_architecture}) that enables users to specify every detail of an experiment, i.e. the datasets, the execution mode (training or HPO mode), the KGE model along with its hyper-parameter values, the details of the evaluation procedure, the seed for the random generator, and the preferred training device (graphics processing unit (GPU) or CPU). To address experts and inexperienced users, experiments can be either configured through the interactive command line interface (CLI) that assists inexperienced users, or programmatically. The CLI ensures that an experiment is configured correctly. In case that users provide an incorrect value for a hyper-parameter such as a negative number for the embedding dimension, the CLI notifies the users and provides an example of a correct input. 

\paragraph{\textbf{Training of KGE Models.}} In PyKEEN we have clearly defined training procedures: KGE models are trained based on the \textit{open world assumption} i.e. triples that are not contained in a KG are not considered as non-existing, but as unknowns which might be true or false facts. The models are trained according the algorithm described by Bordes \textit{et al.}~\cite{bordes2013translating}, and the margin ranking loss and the binary cross entropy are used as loss functions~\cite{nickel2016review}.
Selecting suitable hyper-parameter values is fundamental for the model performance and strongly depends on the expertise and experience of the users. To address both, experienced and inexperienced users, we have developed the \textit{training} and \textit{hyper-parameter optimization mode (HPO)}. In training mode users provide for each hyper-parameter the corresponding value. Optionally, a trained KGE model can be evaluated in training mode. In HPO mode, users have to define for each hyper-parameter a set of possible values (or single values) and PyKEEN assists users to find suitable hyper-parameter values by applying \textit{random search}~\cite{goodfellow2016deep}. The hyper-parameters obtained by the hyper-parameter optimization can be used later to train the final model in training mode.

\paragraph{\textbf{Evaluation of KGE Models.}} Within PyKEEN all the KGE models are evaluated based on the procedure described in Bordes \textit{et al.} ~\cite{bordes2013translating} and the widely applied metrics \textit{mean rank} and \textit{hits@k} are computed~\cite{bordes2013translating}.
Users can provide a set of test triples, or they can use PyKEEN to automatically split the input KG into training and test triples based on a user defined splitting ratio. This is especially relevant if a separate test set is not available. Furthermore, users can specify whether they want to compute the mean rank and hits@k in the \textit{raw} or \textit{filtered setting}. In the filtered setting, artificially created negative samples that are contained as positive examples in the training set will be removed~\cite{bordes2013translating}. Usually, results for both settings are reported.

\paragraph{\textbf{Exporting Experimental Artifacts.}} To ensure the reproducibility of a KGE experiment, we export all relevant experimental artifacts after an experiment is conducted. Specifically, we export a configuration file (JSON) describing the experimental setup, the evaluation results (as JSON file), mappings of entities and relations to unique IDs (JSONs), mappings of entities and relations to their learned embeddings (JSONs), and the trained model in a serialized format (pickle). The exported artifacts can be distributed by our model zoo.

\paragraph{\textbf{Inference.}} Inference can be performed in two ways within PyKEEN. On the one hand, a trained KGE model can be used to provide predictions for a set of triples by calling its \textit{predict} function. On the other hand, we have implemented an inference workflow that provides additional functionalities: for a set of user defined entities and relations, automatically all triple-permutations are created for which predictions are computed. The set of generated triples can be filtered by providing triples that should be removed. This is for instance relevant in a setting, in which predictions for all possible triples except those contained in the training set should be computed. Furthermore, it can be defined that all reflexive triples of the form \begin{math}(e,r,e)\end{math} should be excluded. The output of the inference workflow is a file containing the triples and their predicted scores where the most plausible triples are located at the beginning of the file. 

\subsection{BioKEEN}\label{BioKEEN}
With the development of BioKEEN we demonstrate how KGE research can be transferred to research domains outside the machine learning community. While developing BioKEEN we took into account that expertise in KGE models and in their implementation might be limited in the bioinformatics community. Within BioKEEN we provide direct access to numerous biomedical databases without requiring the user to process them.  

\begin{figure}[t]
\centering
\includegraphics[width=0.9\textwidth]{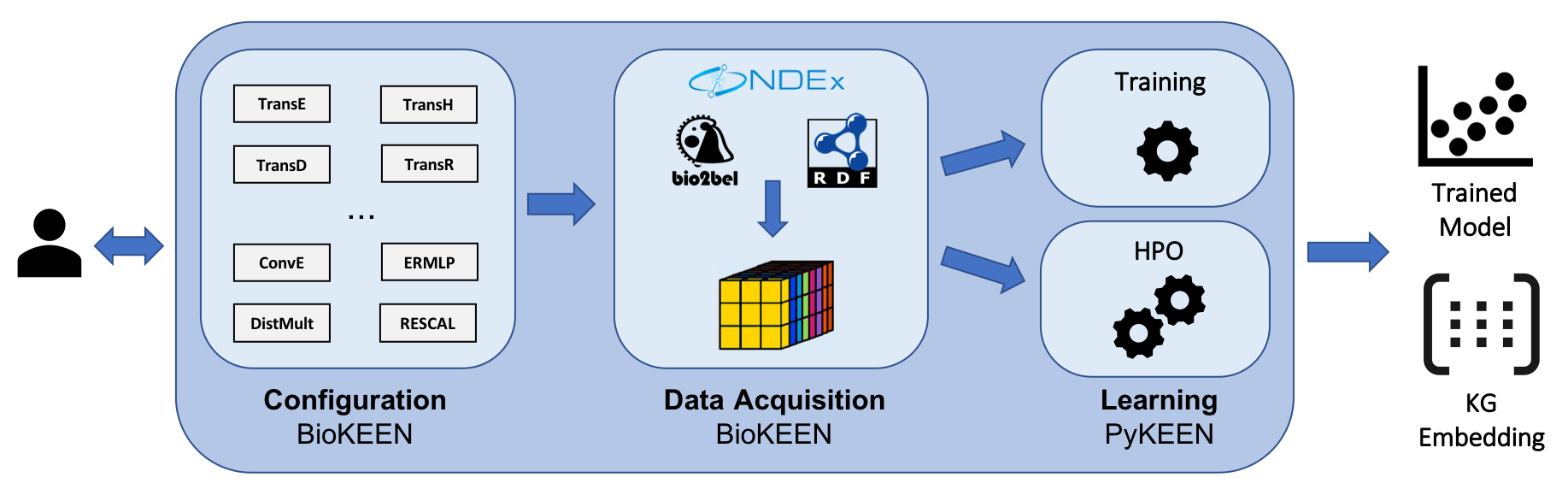}
\caption{BioKEEN's Software Architecture~\cite{ali2019biokeen}.} 
\label{fig:biokeen_software_architecture}
\end{figure}

\paragraph{\textbf{Software Architecture.}}
BioKEEN consists of a three-layered architecture (Figure \ref{fig:biokeen_software_architecture}). Its \textit{configuration layer} is an extension of PyKEEN's configuration layer and enables users to select one of the biomedical databases that are directly accessible through BioKEEN, the \textit{Data Acquisition Layer} provides access to these databases and the learning layer (part of PyKEEN) performs the training of the KGE models.

\paragraph{\textbf{Easy Access to Numerous Biomedical Databases.}}
Within the biomedical domain, numerous databases containing structured knowledge are available~\cite{belleau2008bio2rdf}. However, data pre-processing is a time consuming process. For this reason, we have created the \textit{Data Acquisition Layer} that automatically retrieves and converts the content of numerous biomedical databases and makes it available within BioKEEN (a full list is available at \url{https://biokeen.readthedocs.io/en/latest/bio2bel\_repositories.html}). The data acquisition layer makes use of the Bio2BEL\cite{hoyt2019integration} software to access the databases. Bio2BEL is a framework that gathers biological data sources and represents them in the Biological Expression Language (BEL)\footnote{http://openbel.org/}. By integrating the Bio2BEL software users have direct access to several biomedical databases, can automatically update the database version, and retrieve further databases as they are integrated to Bio2BEL. This functionality allows bioinformaticians to focus on their experiments instead of data pre-processing. 

Overall, the data acquisition layer, the HPO mode and the interactive command line interface are essential features to make KGE research transferable to the domain of bioinformatics considering that the expertise in KGE models and their implementation might be limited. 

\subsection{KEEN Model Zoo}\label{Model-Zoo}
We have created the KEEN Model Zoo as a GitHub project to provide a platform on which researchers can share their experimental artifacts (i.e. trained KGE models, configuration files, evaluation summaries, etc.) that have been created using components of the KEEN Universe. Providing these artifacts publicly will improve the reproducibility of KGE research, and we aim the community to contribute to this project.

To ensure the quality of the model zoo, we have defined following requirements:
i.) conducted experiments must be reported in a scientific paper, ii.) all experimental artifacts that have been created by Py/BioKEEN for an experiment needs to be provided, iii.) the used datasets have to be publicly accessible, iv.) a description of the experiment must be provided, v.) a unit test needs to be implemented checking that the provided model can be instantiated. Within the model zoo, we split experiments based on their research domains (e.g. bioinformatics, scholarly metadata research, etc.), and within each research domain, the experiments are categorized according to the datasets on which the experiments have been conducted. 

Researchers that want to share their experimental artifacts are asked to create a \textit{pull request} that will be reviewed and \textit{merged} into the \textit{master branch} if all requirements are fulfilled. 

\section{Implementation}\label{Implementation}
We have implemented PyKEEN and BioKEEN in Python since it is an established programming language for implementing machine learning models\footnote{https://github.blog/2019-01-24-the-state-of-the-octoverse-machine-learning}. PyTorch~\cite{paszke2017automatic} has been used as the underlying machine learning framework, because it provides flexibility in implementing machine learning models, is easy to debug and through it's GPU support the training procedure can be accelerated. Furthermore, we make use of the scientific Python stack for scientific computing (NumPy\footnote{http://www.numpy.org/}, SciPy\footnote{https://www.scipy.org/}, Scikit-Learn\footnote{https://scikit-learn.org/stable/}, Pandas\footnote{https://pandas.pydata.org/}). Moreover, we apply following community standards: \textit{flake8}\footnote{http://ake8.pycqa.org/en/latest/} to ensure code quality, \textit{setuptools}\footnote{https://github.com/pypa/setuptools/tree/master/setuptools} to create distributions, \textit{pyroma}\footnote{https://github.com/regebro/pyroma} to ensure package metadata standards, \textit{sphinx}\footnote{http://www.sphinx-doc.org/en/master/} to build our documentation and \textit{Read the Docs}\footnote{https://readthedocs.org/} to host it. Finally, \textit{Travis-CI}\footnote{https://travis-ci.org/} is used as continuous integration server.

\textbf{Extensibility.} The KEEN Universe can be extended in various ways. New machine learning related components can be added (extension of PyKEEN is required), further data reader can be implemented to load additional data formats (extension of PyKEEN), further components specifically relevant for the bioinformatics community can be integrated (extension of BioKEEN is required), finally extensions of PyKEEN specialized for further research domains can be created. Here, we describe how new machine learning components can be integrated into our ecosystem by extending PyKEEN. Figure \ref{fig:pykeen_modules} depicts the sub-modules of PyKEEN and the most relevant with regards to an extension are the \textit{KGEModel} and the \textit{HPOOptimizer} modules. The modular architecture of PyKEEN facilitates its extension.

\textbf{Integration of an Additional KGE Model.} Within PyKEEN, a \textit{KGEModel} module interacts with the \textit{Pipeline}, the \textit{Training}, and the \textit{Inference} module (Figure \ref{fig:pykeen_modules}). To ensure that a new KGE model can interact with these modules, it needs to provide implementations of a \textit{forward()} and a \textit{predict()} function. The \textit{forward} should expect two multi-dimensional arrays (tensors) containing the batch of positive and negative training triples (or a batch of training triples and corresponding labels; depends on the KGE model) and return the loss value computed for this batch. The \textit{predict} function should expect a tensor of triples for which predictions should be computed and returned. There are no further constraints for the model implementation.

\textbf{Integration of an Additional Hyper-Parameter Optimization Algorithm.} Currently, random search is applied to perform hyper-parameter optimizations and \textit{RandomSearchHPO} is the corresponding module. It extends our abstract class \textit{AbstractHPOoptimizer} which contains the two abstract functions \textit{optimise\_hyperparams} and \textit{sample\_parameter\_value}, where the former is used to initiate the optimization procedure and the latter is called in each optimization iteration to sample new hyper-parameter values. To add a new hyper-parameter optimizer, the respective module has to extend the abstract class \textit{AbstractHPOoptimizer} and provide implementations for its two abstract functions to ensure that the optimizer can interact with the \textit{Pipeline}, the \textit{Training}, and the \textit{Evaluator} module.

Overall, the modular architecture of PyKEEN and the simple API of the KGE and hyper-parameter optimization modules facilitate the integration of new machine learning components to PyKEEN.

\section{Availability and Maintenance}
\paragraph{\textbf{Availability.}} PyKEEN, BioKEEN and the KEEN Model Zoo are available at our GitHub repositories under the MIT License. Furthermore, PyKEEN and BioKEEN are also available through PyPI enabling users to install the software packages easily through \textbf{pip}. 

\paragraph{\textbf{Maintenance.}} We aim that researchers from different communities (e.g., semantic web, machine learning, bioinformatics, crop science) will support us in maintaining and extending the KEEN Universe. Before this state is reached, the maintenance of the KEEN Universe is ensured through the Bio2Vec\footnote{http://bio2vec.net/} and the German national funded BmBF project MLwin\footnote{https://mlwin.de/} at least till 2022.

\section{Evaluation of the Usability of the KEEN Universe}
Usability is defined as the extent a software system can be used to achieve a goal with \textit{effectiveness} (extent to which the tasks can be completed), \textit{efficiency} (resources required to achieve the goals) and \textit{satisfaction} (feeling of the users towards the software) in a specified context~\cite{abran2003usability}. We evaluate these aspects based on two case scenarios: co-author recommendations for a scholarly KG, and the predictions of crosstalks and hierarchies between biological pathways.

\subsection{Co-Author Recommendations Based on KGEs}
In the work of Henk \textit{et al.}~\cite{KGErecommender}, PyKEEN has been used to provide co-author recommendations based on KGEs for a scholarly KG. The KG contains the entity types \textit{author}, \textit{paper}, \textit{department} and \textit{event}. Furthermore, it contains the relationship types \textit{isAuthorOf}, \textit{isCoAuthorOf}, \textit{isAffiliatedIn} and \textit{isPublished}. The goal was to evaluate co-author recommendations i.e. triples of the form (\textit{author}, \textit{isCoAuthorOf}, \textit{author}). For additional information including the experimental set-up and the evaluation, we refer to ~\cite{KGErecommender} and the final experimental artifacts are available at our model zoo (\sloppy \url{https://github.com/SmartDataAnalytics/KEEN-Model-Zoo/tree/master/scholarly_data_related_recommendations/SG4MR/sg4mr_model_01}).

\paragraph{\textbf{Effectiveness.}} The KEEN Universe provides all components to completely achieve the goal: PyKEEN has been used to train four KGE models (DistMult, TransE, TransH and TransR) on the KG, and through the hyper-parameter optimization mode, a suitable combination of KGE model and hyper-parameter values has been automatically determined. Based on the model that performed best, we have used the inference workflow to provide co-author recommendations which have been manually evaluated by a domain expert that classified the top predictions as valid recommendations. 

\paragraph{\textbf{Efficiency.}} Considering efficiency with regards to the computation time, we made use of the GPU support of PyKEEN (PyTorch) to reduce the training time. The models have been trained on a single GPU. Efficiency with respect to the time necessary to learn the software to be able to solve the task, the main author could quickly set-up and run her experiments through the command line interface which assisted and ensured that the experiments have been configured correctly. The whole process has been performed without any programming required by the author.

\paragraph{\textbf{Satisfaction.}} The main author didn't have any prior knowledge about KGEs and the software ecosystem, but she could easily achieve her goals. This positive experience has helped her to get into the field of KGEs.

\subsection{Prediction of Cross-talks and Hierarchies Between Biological Pathways}
In the work of Ali \textit{et al.}~\cite{ali2019biokeen}, BioKEEN has been used to predict novel cross-talks and hierarchies between biological pathways. ComPath~\cite{domingo2018compath}, a novel database for biological pathways has been used to train the KGE models. ComPath contains two types of relationships: \textit{equivalentTo} expressing that two pathways correspond to the same biological process, and \textit{isPartOf} expressing a hierarchy of pathways. Again, we refer to~\cite{ali2019biokeen} for additional information and to \url{https://github.com/SmartDataAnalytics/KEEN-Model-Zoo/tree/master/bioinformatics/ComPath/compath_model_01} to access the experimental artifacts of the final model.

\paragraph{\textbf{Effectiveness.}} The KEEN Universe provides all components to completely achieve the goal: We have used BioKEEN to train five KGE models (UM, DistMult, TransE, TransH and TransR) on ComPath that is directly accessible through BioKEEN. We performed a hyper-parameter optimization to find the best combination of KGE model and hyper-parameters, showed the sensibility of choosing appropriate hyper-parameter values and the effectiveness of the HPO mode to find suitable hyper-parameter values (performance increase from 19.10\% to 63.20\% for the hits@k metric). The final model has been used to predict new interactions between pathways based on the inference workflow. The top predictions have been evaluated by domain experts and we found following novel links that have been added to ComPath: the first link states that the \textit{TGF-beta signaling pathway} is equivalent to the \textit{TGF-beta Receptor Signaling} pathway, and the second link expresses that \textit{Lipoic Acid} is part of \textit{Lipid Metabolism}. 

\paragraph{\textbf{Efficiency.}} Because ComPath is not a large KG, we trained the KGE models on a single CPU (efficiency regarding computation time). Furthermore, no pre-processing of the dataset was required since it is directly accessible within BioKEEN. Although the primary author has no domain expertise regarding pathway interactions, he effortlessly provided new predictions to domain experts who validated them (efficiency with respect to use the software for solving the task). 

\paragraph{\textbf{Satisfaction.}} Through BioKEEN the main author was able to get to know a new application area in the field of bioinformatics. Furthermore, researchers from different research fields could work successfully in an interdisciplinary team.

\section{Related Work}
\paragraph{\textbf{Supported KGE Models.}} KGE models can be divided into \textit{translational distance models (TDM)} and \textit{semantic matching models (SMM)} where the former compute the plausibility of a fact by a distance function (e.g. using the  Euclidean norm) and the latter apply similarity-based scoring functions (considering the similarity of the latent features of the entities and relations)~\cite{wang2017knowledge}. Table \ref{tab:supported_kge_models} lists all the KGE models that are currently available within the KEEN Universe.

\begin{table}[t]
\centering
\caption{KGE Models available within the KEEN Universe.}\label{tab:supported_kge_models}
\begin{tabular}{|l|l|l|}
\hline
Type & Reference & Model \\
\hline
\multirow{6}{*}{TDM} & ~\cite{wang2017knowledge} & TransE \\
& ~\cite{wang2017knowledge} & TransH \\
& ~\cite{wang2017knowledge} & TransR \\
& ~\cite{wang2017knowledge} & TransD  \\
& ~\cite{wang2017knowledge} & Unstructured Model (UM) \\
& ~\cite{wang2017knowledge} & Structured Embedding (SE) \\
\hline
\multirow{4}{*}{SMM} &~\cite{nickel2011three} & RESCAL \\
& ~\cite{wang2017knowledge} & DistMult \\
& ~\cite{wang2017knowledge} & ERMLP \\
& ~\cite{dettmers2017convolutional} & ConvE \\
\hline
\end{tabular}
\end{table}

\paragraph{\textbf{Existing Ecosystems for KGE Models.}} The available software for KGE models is limited, and an ecosystem like the KEEN Universe is to the best of our knowledge unique. However, there exist software projects that provide implementations of different KGE models. One of them is scikit-kge\footnote{https://github.com/mnick/scikit-kge} that provides implementations of three KGE models and different negative sampling approaches. The project doesn't seem to be maintained since the last commit dates back to the year 2016. A recently published framework which enables users to train and evaluate several KGE models is OpenKE~\cite{han2018openke} that can be compared to PyKEEN (Section~\ref{PyKEEN}). While allowing users to reproduce KGE experiments, we argue that it has not been developed with the goal of making KGE research transferable to domains outside the machine learning community and usable for both, experts and non-experts. For instance, it supports only one data format (a text-file consisting of three columns) whereas within PyKEEN a KG can be provided as tab separated values, RDF and from NDEx (Section \ref{supported_data_formats}). Users without expertise in programming might face difficulties to run the software since it doesn't provide an interactive command line interface, and users without expertise in KGE models might have issues in finding appropriate combinations of KGE models and corresponding hyper-parameter values since it doesn't provide a hyper-parameter optimization procedure. Further software repositories containing implementation for different KGE models can be found at\footnote{https://github.com/bookmanhan/Embedding} and \footnote{https://github.com/TimDettmers/ConvE}.

\section{Limitations and Future Work}
Currently, all the KGE models available within the KEEN Universe make only use of the triples of a KG. However, several KGs contain additional information such as textual descriptions of entities, images and numerical values which can be used to train multimodal KGE models. Based on multimodal data, KGE models can be developed that are capable of performing inference among different KGs which is currently not possible with models that are trained only based on the entities and relations of a KG~\cite{xie2016representation}. We plan to integrate an additional software package to our ecosystem that contains implementations of multimodal KGE models.

Within PyKEEN, negative samples are created based on the approach described in Bordes \textit{et al.}~\cite{bordes2013translating}. However, it has been shown that alternative approaches such as \textit{bern}~\cite{wang2014knowledge} can yield better performance. Therefore, we aim to implement additional negative sampling approaches.

KGE models are evaluated within our ecosystem based on the widely applied metrics \textit{mean rank} and \textit{hits@k}, but additional metrics such as the \textit{AUC-ROC} and \textit{AUC-PR} curve might be of interest~\cite{nickel2016review}. Furthermore, Sharma \textit{et al.}~\cite{sharma2018towards} propose a geometrical analysis of learned embeddings that can provide valuable insights. We plan to implement these additional evaluation metrics within the KEEN Universe.

\paragraph{Acknowledgments}
This work was partly supported by the KAUST project grant Bio2Vec (grant no.~3454), the European Union’s Horizon 2020 funded project BigDataOcean (GA no.~732310), the CLEOPATRA project (GA no.~812997), and the German national funded BmBF project MLwin.

\bibliographystyle{splncs04}

\end{document}